\documentclass[lettersize,journal]{IEEEtran}
\usepackage{amsmath,amsfonts}
\usepackage{algorithmic}
\usepackage{algorithm}
\usepackage{array}
\usepackage{xcolor}
\usepackage{textcomp}
\usepackage{stfloats}
\usepackage{url}
\usepackage{hyperref}
\usepackage{verbatim}
\usepackage{graphicx}
\usepackage{cite}
\usepackage{booktabs}
\usepackage{adjustbox}
\usepackage{multirow}
\usepackage{subcaption}
\usepackage{makecell}
\usepackage{soul}

\usepackage{xcolor}
\usepackage{colortbl}
\newcolumntype{C}[1]{>{\centering\arraybackslash}m{#1}}

\usepackage[capitalize]{cleveref}
\crefname{section}{Sec.}{Secs.}
\Crefname{section}{Section}{Sections}
\Crefname{table}{Table}{Tables}
\crefname{table}{Tab.}{Tabs.}

\begin{document}

\title{From Offline to Periodic Adaptation for Pose-Based Shoplifting Detection in Real-world Retail Security}
\author{Shanle Yao\IEEEauthorrefmark{1},~\IEEEmembership{Student Member,~IEEE,}
        Narges Rashvand\IEEEauthorrefmark{1},~\IEEEmembership{Student Member,~IEEE,}
        Armin Danesh Pazho\IEEEauthorrefmark{1},~\IEEEmembership{Student Member,~IEEE,}
        Hamed Tabkhi,~\IEEEmembership{Member,~IEEE}
\thanks{The authors are with the Electrical and Computer Engineering Department, The University of North Carolina at Charlotte, Charlotte,
	NC, 28223 USA.\\
	\{syao, nrashvan, adaneshp,  htabkhiv\}@charlotte.edu\\
	\IEEEauthorrefmark{1} Corresponding authors have equal contribution.}}

\markboth{IEEE INTERNET OF THINGS JOURNAL}%
{Shanle Yao \MakeLowercase{\textit{et al.}}: Title: To Be Determined}

\IEEEpubid{0000--0000/00\$00.00~\copyright~2026 IEEE}

\maketitle

\begin{abstract}
Shoplifting is a growing operational and economic challenge for retailers, with incidents rising and losses increasing despite extensive video surveillance. Continuous human monitoring is infeasible, motivating automated, privacy-preserving, and resource-aware detection solutions. In this paper, we cast shoplifting detection as a pose-based, unsupervised video anomaly detection problem and introduce a periodic adaptation framework designed for on-site Internet of Things (IoT) deployment. Our approach enables edge devices in smart retail environments to adapt from streaming, unlabeled data, supporting scalable and low-latency anomaly detection across distributed camera networks. To support reproducibility, we introduce \textit{RetailS}, a new large-scale real-world shoplifting dataset collected from a retail store under multi-day, multi-camera conditions, capturing unbiased shoplifting behavior in realistic IoT settings. For deployable operation, thresholds are selected using both F1 and \(H_{\text{PRS}}\) scores, the harmonic mean of precision, recall, and specificity, during data filtering and training. In periodic adaptation experiments, our framework consistently outperformed offline baselines on AUC-ROC and AUC-PR in 91.6\% of evaluations, with each training update completing in under 30 minutes on edge-grade hardware, demonstrating the feasibility and reliability of our solution for IoT-enabled smart retail deployment.
\end{abstract}

\section{Introduction}
\label{sec:intro}

Shoplifting has become a pressing operational and economic problem for retailers, straining profit margins, disrupting store operations, and weakening community trust. The scale of shoplifting is underscored by alarming statistics. Reported shoplifting incidents increased 15\% year over year, rising from 1 million in 2022 to 1.15 million in 2023 \cite{capitaloneshopping2025}. Compared to 2019, annual shoplifting incidents in 2023 rose by 93\%\cite{capitaloneshopping2025}. The financial impact is equally severe: retailers lost an estimated 45 billion USD in 2024, with losses projected to exceed 53 billion USD by 2027 \cite{freedomforallamericans,capitaloneshopping2025}. Yet despite this scale, stores apprehend shoplifters in only about 2\% of cases \cite{capitaloneshopping2025,Rashvand_2025_WACV}. Taken together, these trends reveal shoplifting as not just an operational challenge but a nationwide socio-economic issue, affecting small neighborhood stores to large retail chains, and creating urgency for technological countermeasures. \footnote{Dataset repository: \url{https://github.com/TeCSAR-UNCC/RetailS}}

Retailers have invested heavily in video surveillance to address this growing threat, but the sheer volume of footage makes continuous human monitoring infeasible. This gap highlights the need for \textbf{automated, IoT-enabled video analytics systems} capable of running across distributed edge devices. IoT-based smart retail infrastructure, powered by networked cameras and embedded computing platforms, enables real-time monitoring, localized decision making, and privacy-preserving analytics at scale. Recent studies in urban safety and traffic monitoring \cite{kefayat2025urban,arroyo2015expert, sun2019deep, rahmanidehkordi2024traffic,rahmanidehkordi2024enhancing, babaey2025detecting,rashid2024quantifying, ghorbani2025examining} demonstrate the potential of IoT-driven AI for security-critical domains. In this context, Video Anomaly Detection (VAD) provides a structured framework to learn behavioral norms and detect deviations, making it well-suited for IoT-enabled smart retail systems.  

\begin{figure}[]
    \centering
    \includegraphics[clip,trim={40 510 10 0},width=1\columnwidth]{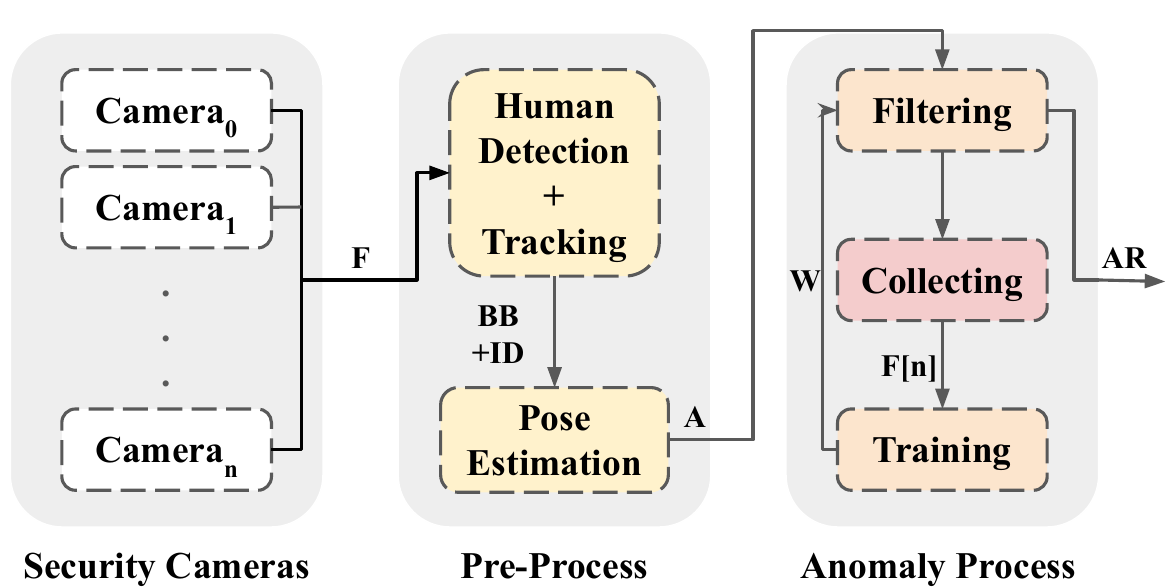}
    \caption{A conceptual overview of an IoT-enabled shoplifting detection system with continual unsupervised anomaly detection. Frames (F) from distributed surveillance cameras are preprocessed at the edge to produce annotations (A), including bounding boxes (BB), track IDs (ID), and human poses. The anomaly detector filters frames for real-time screening and outputs anomaly results (AR). A collector accumulates pseudo-filtered frames F[n] for continual unsupervised adaptation. After each training cycle, updated model weights (W) are pushed back to the edge for the next cycle of detection.}
    \label{fig:intro}
\end{figure}

\IEEEpubidadjcol
In human-centric VAD, prior work has explored multiple representations for modeling human motion, including RGB pixels, pose keypoints, and heatmaps. Pixel-based approaches are highly expressive, but their deployment in IoT-driven retail environments is constrained by limitations: privacy concerns associated with appearance-based monitoring; computational and memory constraints on edge devices; and sensitivity to environmental variability, such as illumination changes, and layouts, which undermines robust real-time inference. Heatmap-based representations partially reduce privacy risks by discarding raw information and encoding joints; however, it remains dense, image-like tensors, brings computational overhead that limits scalability in large-scale deployments \cite{maldonado2025adversarially, wang2019deep, luo2021rethinking}. In contrast, pose-based representations operate directly on skeletal sequences, offering a compact, privacy-preserving data of human motion that is robust to visual noise and suited for IoT environments, while 3D pose detection brings computation overhead trade off with 2D pose-based representation. \cite{zhang2025wivipose}

Importantly, recent study demonstrates that adopting 2D pose-based representations does not entail a loss in detection capability. On the SHT dataset \cite{liu2018future}, state-of-the-art pixel-based methods—including SSMTL++v2 \cite{barbalau2023ssmtl++}, Jigsaw-VAD \cite{wang2022video}, and AnomalyRuler \cite{yang2024follow}—achieve AUC-ROC scores of 83.80, 84.30, and 85.20, respectively. Meanwhile, pose-based approaches such as MoPRL \cite{yu2023regularity}, STG-NF \cite{hirschorn2023normalizing}, and SPARTA \cite{noghre2024human} attain competitive performance, with AUC-ROC scores of 83.35, 85.90, and 85.75. These results show that pose-based representations preserve discriminative power while simultaneously offering decisive advantages in privacy protection, computational efficiency, and deployment feasibility.
However, pose-based approaches remain under-studied and constrained by the lack of realistic, large-scale datasets designed for IoT-driven streaming and multi-camera scenarios.  

To address these challenges, we frame shoplifting detection as a pose-based, unsupervised anomaly detection problem with periodic adaptation , designed for real-world IoT retail deployment. Unlike prior work centered on offline evaluation or limited dataset, our approach targets long-term operation, where labeled anomalies are scarce, behaviors drift over time, and retraining must run efficiently on local hardware. Based on this framing, we design an end-to-end IoT-oriented pipeline (\cref{fig:intro}) that combines anonymized pose extraction, unsupervised anomaly modeling, and periodic adaptation. While pose representations, unsupervised VAD, and periodic training have been studied independently, their integrated design and evaluation under real retail conditions remains limited. To support deployment, we introduce pseudo-data filtering, size-aware retraining schedules, and local updates. Supporting this pipeline, we collaborated with a local retailer to collect a significantly large dataset, \textit{RetailS}, containing both normal shopping activities and shoplifting incidents, captured across multiple camera views in real-world IoT environments. In addition, we introduce a new staged shoplifting branch, where controlled shoplifting scenarios were systematically recorded to cover diverse concealment strategies. Leveraging this setup, we conduct extensive experiments with our proposed framework, achieving superior AUC-ROC and AUC-PR in 91.6\% of evaluations, with each training update completing in under 30 minutes, demonstrating both feasibility and reliability for IoT-enabled retail deployment.  

In summary, this study makes the following contributions:
\begin{itemize}
  \item We introduce a pose-based periodic adaptation pipeline with pseudo-decision filtering for IoT-enabled shoplifting detection.  
  \item We present \textit{RetailS}, a large-scale, multi-camera, pose-based dataset of both staged and real-world shoplifting incidents, designed for IoT streaming analysis.  
  \item We provide a comprehensive benchmark of state-of-the-art VAD models under offline and online IoT deployment conditions, including time-efficiency analyses, demonstrating the practical benefits of our approach for smart retail environments.  
\end{itemize}

\section{Related Works}

Research in shoplifting detection has evolved along two main directions: dataset development and algorithmic innovation. Several datasets have been introduced \cite{ansari2022expert, arroyo2015expert, muneer2023shoplifting}, providing video recordings of shoplifting-related behaviors. However, these datasets are typically staged, limited in diversity, and often restricted to single-camera perspectives. Even the widely used UCF-Crime dataset \cite{sultani2018real}, while containing real-world surveillance footage including shoplifting, was not specifically designed for retail security and lacks detailed annotations of shoplifting-specific actions. More recently, PoseLift \cite{Rashvand_2025_WACV} introduced pose sequences captured in retail environments, enabling human-centric and privacy-preserving analysis beyond raw pixels. Yet, most existing datasets remain insufficient for studying adaptive and resource-constrained deployment challenges that are central to IoT systems.  

On the algorithmic front, several studies have proposed deep learning methods for shoplifting detection. Pixel-level approaches \cite{kirichenko2022detection,nazir2023suspicious,ansari2022expert} commonly build on CNN-RNN hybrids, object detection with tracking, or attention-based feature fusion, primarily evaluated on UCF-Crime. Other pixel-based anomaly detection methods \cite{huang2023multi, zaheer2022generative, georgescu2021background, georgescu2021anomaly, barbalau2023ssmtl++, wang2022video, chen2021nm, RL00, Gaus_2023vand, Lappas_2024vand7} extend these ideas with generative models or self-supervised learning. While effective in offline benchmarks, such methods are compute-heavy, sensitive to environmental variations, and raise privacy concerns in IoT deployment contexts.  

Pose-based frameworks \cite{yu2023regularity, chen2023multiscale, jain2021posecvae, Heckler_2023vand3, Baradaran_2023vand4, zeng2021hierarchical, huang2022hierarchical, rodrigues2020multi, noghre2024exploratory, morais2019learning} have emerged as a promising alternative, shifting the focus from raw pixels to skeletal dynamics. Building on this idea, recent methods such as GEPC \cite{markovitz2020graph}, STG-NF \cite{hirschorn2023normalizing}, TSGAD \cite{noghre2024exploratory}, Shopformer \cite{rashvand2025shopformer}, and SPARTA \cite{noghre2024human} apply spatio-temporal graphs, normalizing flows, Variational Auto Encoders (VAEs), and transformers for unsupervised behavior modeling. Pose-based anomaly detection offers strong privacy benefits and robustness, making it attractive for IoT-enabled retail systems where resource constraints and demographic biases must be carefully addressed. However, these methods are largely designed for static evaluation and have not been examined in the context of distributed, streaming IoT deployments.  

The notion of continual anomaly detection has also been studied inconsistently. Many works equate it with real-time inference using fixed models \cite{zhang2023online,karim2024real,Yang_2024vand5}, whereas others explore online learning with incremental updates \cite{Doshi_2020_CVPR_Workshops,doshi2021online,doshi2023towards}. Recent studies \cite{yao2024evaluating} highlight that static decision boundaries quickly degrade in dynamic environments, underscoring the need for adaptive learning. While cross-domain generalization has been tested by evaluating models on unseen datasets \cite{doshi2023towards,danesh2023chad,Yang_2023vand2}, explicit mechanisms for domain adaptation in streaming IoT contexts remain underexplored.  

Despite these advances, a critical limitation persists: most shoplifting detection methods are designed and validated in offline laboratory conditions, with little attention to the algorithmic and learning challenges of real-world IoT deployment. Specifically, they (1) assume static thresholds and fixed operating conditions, (2) overlook resource constraints of edge-grade devices in IoT environments, and (3) ignore the need for continual adaptation to seasonal, environmental, or behavioral drift in retail stores. As a result, such methods often suffer high false-positive rates and limited scalability when deployed in practice.  

In contrast, IoT-driven smart retail environments demand models that can operate reliably across distributed camera networks, adapt to non-stationary data streams, and update in near real-time without requiring central retraining. Continual learning has been applied in other IoT domains such as autonomous driving and human activity recognition, but its application to shoplifting detection remains largely unexplored. Addressing this gap requires a shift from benchmark-driven evaluation to frameworks explicitly designed for on-device learning, adaptation, and deployment within real-world IoT systems.
\section{Methodology} 
\label{sec:method}

\begin{figure}[]
    \centering
    \includegraphics[clip,trim={20 470 20 0},width=1\columnwidth]{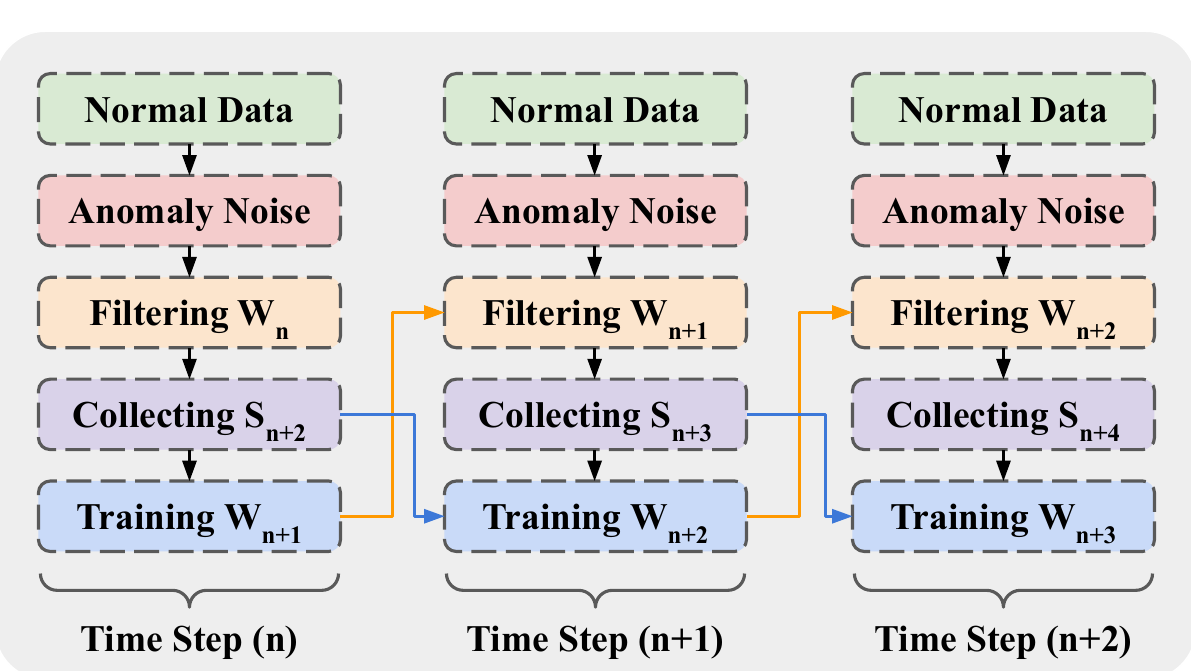}
    \caption{A conceptual overview of our IoT-oriented continual unsupervised anomaly detection pipeline with pseudo filtering, collection, and training. Model updates are designed to run incrementally on edge-grade devices, enabling scalable deployment across distributed surveillance nodes.}
    \label{fig:concept}
\end{figure}

To realize a practical, IoT-deployable framework for continual, unsupervised shoplifting detection with pose-based models, we design a three-stage pipeline that mirrors real-world smart retail operations: \textbf{filtering}, \textbf{collection}, and \textbf{training}. This pipeline is shown in \cref{fig:concept}. In the filtering stage, a pretrained model evaluates the incoming multi-camera video stream, generating anomaly scores and forwarding pseudo-labeled normal frames using adaptive thresholds. The collection stage aggregates these filtered frames into a buffered dataset under resource-aware scheduling. Once a buffer reaches a preset size, the training stage fine-tunes the model to the local retail environment. Updated weights are then pushed back to the filtering service, with a fixed two-step lag inherent to the pipeline. This design ensures continuous adaptation to evolving behaviors while meeting the compute and latency constraints typical of IoT deployments.  

\subsection{Input Data}
As noted in \cref{sec:intro}, reliable anomaly detection in IoT-enabled retail depends on robust per-person detection, tracking, and pose extraction from live multi-camera streams. These signals are inherently sensitive to IoT-specific challenges, including heterogeneous camera hardware, variable bandwidth, and streaming conditions such as location, angle, and field of view. Consequently, anomaly definitions become domain- and deployment-specific, requiring periodic adaptation.  

To emulate this variability, we collected a new multi-day, multi-camera pose-based dataset in a live retail location (detailed in \cref{sec:dataset}). For training, we begin with normal shopping streams and inject staged anomaly frames at a 9:1 ratio, reflecting the rarity of theft. 


\subsection{Filtering}
Pose-based anomaly detectors output uncalibrated scores per frame, not probabilities, making threshold selection critical for IoT deployments where false alarms translate into operational overhead. We evaluate two adaptive thresholding methods that balance precision, recall, and robustness against noise:  

\subsubsection{F1 Score}
The F1 score selects the threshold maximizing the harmonic mean of Precision and Recall,
\[
F_{1}=2\,\frac{PR}{P+R},
\]
with \(P=\) Precision and \(R=\) Recall. While widely used in imbalanced domains such as retail video, F1 ignores true negatives and thus does not directly control the false–positive rate (FPR), a major concern for IoT deployments where excessive alerts can overwhelm store staff and network resources.  

\subsubsection{$H_{\text{PRS}}$}
To explicitly address false alarms in real-world IoT environments, we also optimize
\[
H_{\text{PRS}}=\frac{3}{\tfrac{1}{P}+\tfrac{1}{R}+\tfrac{1}{S}}
=\frac{3PRS}{PR+RS+SP},
\]
the harmonic mean of Precision (\(P\)), Recall (\(R\)), and Specificity (\(S=1-\mathrm{FPR}\)). Because the harmonic mean penalizes any weak component, maximizing \(H_{\text{PRS}}\) yields thresholds that balance anomaly sensitivity with practical false alarm control. We calibrate \(\text{thr}_{F1}\) and \(\text{thr}_{H_{\text{PRS}}}\) on a validation split and report evaluation under both operating conditions.  

\subsection{Collection}
A distinctive challenge in IoT-enabled periodic adaptation is the unpredictable yield of usable frames from distributed camera streams due to variable traffic, occlusion, and network bandwidth. To address this, we implement a \textbf{time-sliced, cross-camera collection scheme} that aggregates frames across all cameras before triggering a training update. This prevents skew toward high-activity cameras and ensures more representative sampling of store-wide activity.  

We test two IoT-aligned schedules: (i) half-day windows (\(\approx6\) hours), and (ii) full-day windows (\(\approx12\) hours). These correspond to natural store operational cycles and yield 20 and 10 training updates respectively in our experiments. This approach stabilizes continual updates, improves generalization across the IoT system, and allows predictable compute allocation on resource-constrained edge devices.  

\subsection{Training}
During training, we standardize temporal settings across models to provide fair and efficient comparison under IoT constraints: window size is fixed at 24 frames and stride size at 6. This balances temporal context with responsiveness in the filtering stage while keeping computational requirements suitable for edge-grade processors, and follows common practice in prior work to enable fair comparison with existing methods. Default hyperparameters (learning rate, epochs, etc.) from original repositories are maintained.  

For evaluation, we report both AUC-ROC and AUC-PR. AUC-ROC reflects ranking ability across thresholds but underemphasizes false negatives, while AUC-PR better captures rare-event performance under imbalance, aligning with the needs of IoT anomaly detection where both missed events and nuisance alarms directly affect operational viability. Together, these metrics provide a comprehensive view of detection reliability for IoT-enabled retail deployment.

\subsection{Sensing-Learning Interaction and Feedback Control}

The framework further establishes a structured interaction between edge-level sensing and back-end adaptive learning. The sensing policy acts as a gatekeeper, selectively admitting low-anomaly frames into the training buffer to form a pseudo-labeled normal distribution. This mechanism allows the model to track gradual behavioral and environmental drift without requiring manual annotations. To prevent spatial bias from high-traffic cameras, a cross-camera collection scheme aggregates data store-wide before any model update is triggered, ensuring a balanced and representative training distribution.

To maintain stability and avoid bias amplification within this closed feedback loop, we introduce three control mechanisms. First, fixed anomaly thresholds calibrated offline are used during deployment, preventing recursive relaxation of the normality criteria. Second, each retraining cycle enforces a 9:1 normal-to-abnormal mixing ratio by injecting curated anomaly samples, preserving a stable decision boundary. Finally, model updates are temporally damped through fixed-interval training schedules (half-day or full-day), with parameters pushed to the edge only after a complete update cycle, effectively suppressing rapid oscillations in detection performance.
\section{Dataset}
\label{sec:dataset}

\subsection{Data Collection}
\textit{RetailS} was collected in collaboration with a U.S. retail store that granted research access to its in-store CCTV system. Over 10 consecutive days of normal operating hours, six indoor cameras captured high-angle, multi-view video at $1080\times720$ resolution and 15 FPS. This multi-camera, multi-day design reflects the heterogeneous, streaming nature of IoT-enabled surveillance networks.  

The dataset consists of three complementary parts: (i) a training set of normal customer behavior, (ii) a staged test set where authorized researchers enacted shoplifting under controlled yet realistic retail conditions, and (iii) a real-world test set consisting of actual shoplifting incidents provided by the store’s security team, spanning two years of IoT surveillance logs. Together, these subsets form a balanced, deployment-oriented benchmark for pose-based shoplifting detection under IoT settings.  

\subsubsection{Normal Customer Behavior}
To support both offline benchmarking and continual IoT adaptation, the dataset includes large-scale recordings of normal customer activities. Unlike prior datasets staged in lab environments, these samples were captured from natural retail traffic without intervention, reflecting the variability of day-to-day IoT surveillance streams.  

\subsubsection{Real-World Shoplifting Events}\label{sec:realdata}
The real-world portion captures incidents extracted from the store’s IoT-based surveillance system during regular business hours. This test set expands on \cite{Rashvand_2025_WACV} and provides a more comprehensive and operationally realistic benchmark. Incidents cover diverse concealment strategies, including: (1) hiding items in pants, (2) concealing items in hoodie pockets, (3) placing items in a bag while standing, (4) placing items in a bag on the floor, and (5) hiding items under a jacket. These reflect adversarial behaviors that IoT-enabled detection systems must capture reliably.  

\subsubsection{Staged Shoplifting Events}
Prior datasets \cite{arroyo2015expert,ansari2022expert,muneer2023shoplifting} were often recorded in static lab environments with fixed viewpoints, scripted behaviors, and minimal occlusion, limiting their relevance to IoT deployments. In contrast, our staged set was recorded in the same retail environment as the normal and real-world sets, using multiple cameras and naturalistic pace. Authorized researchers performed concealment behaviors under realistic conditions such as occlusion, crowd traffic, and varied shelf contexts.  

\begin{figure}[]
    \centering
    \includegraphics[clip,trim={550 305 150 500},width=1\columnwidth]{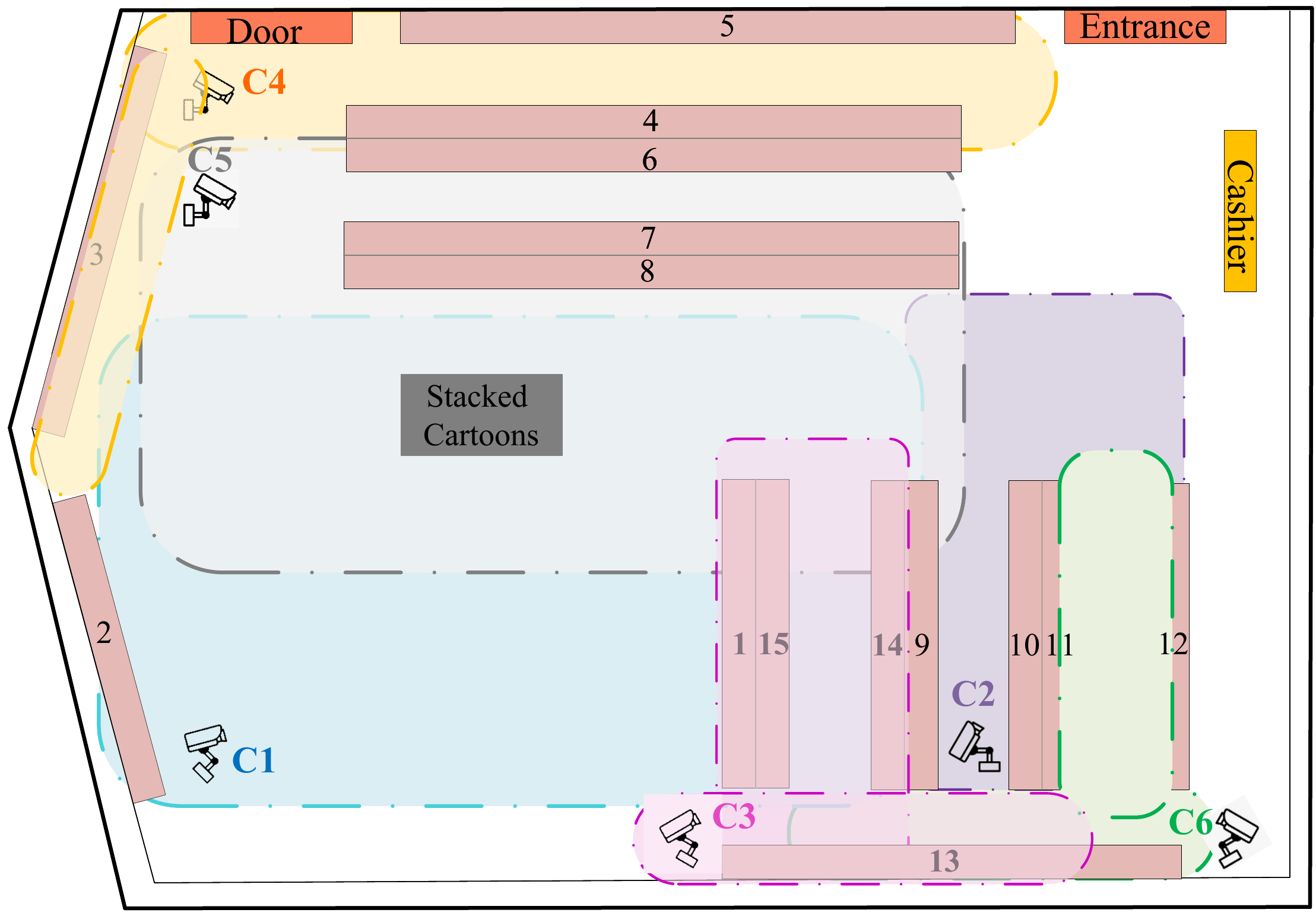}
    \caption{Bird’s-eye view of the retail store in our RetailS dataset, showing six IoT-connected camera locations and their coverage areas.}
    \label{fig:store_layout}
\end{figure}

To further increase spatial diversity for IoT evaluation, staged events were performed in front of 15 different shelves strategically distributed across the store. This ensures coverage of different spatial contexts, enhancing cross-camera variability that is central to IoT streaming environments. The camera locations and their coverage areas for the RetailS dataset are shown in \cref{fig:store_layout}.  

\subsection{Data Preparation and Annotation}
As emphasized in \cref{sec:intro}, IoT-driven deployments must balance accuracy with privacy and fairness. To this end, we avoid raw pixel-level video and instead represent human activity through anonymized pose sequences. This abstraction preserves privacy while retaining behavioral dynamics, aligning with both ethical guidelines and practical IoT regulations.  

All surveillance footage was converted into pose sequences using a multi-stage pipeline for person detection, tracking, and pose estimation, following established practices in prior work. Since the objective of this paper is to study shoplifting detection given pose representations, we adopt a standard pose extraction method \cite{pazho2023ancilia, alinezhad2023understanding, duan2022revisiting}.First, YOLOv8 \cite{yolov8_ultralytics} with ByteTrack \cite{zhang2022bytetrack} generated person bounding boxes with consistent IDs across frames. Then, HRNet \cite{sun2019deep} extracted 17 keypoints per individual in COCO17 format \cite{lin2014microsoft}. This yields anonymized pose streams that capture movement dynamics across time.  

Frame-level annotations were performed manually, labeling shoplifting frames as 1 and normal frames as 0. To improve data quality for IoT-based learning:  
\begin{itemize}
    \item Camera-specific regions of interest (ROIs) were defined, focusing on shelving zones with high visibility and excluding irrelevant areas.  
    \item Keypoints were interpolated where missing, and an 8-frame smoothing filter was applied to reduce noise and ensure stable motion sequences under IoT streaming conditions.  
    \item All annotations were cross-validated by two independent reviewers.  
\end{itemize}  

\begin{figure}[]
    \centering
    \includegraphics[clip,trim={0 0 0 0},width=1\columnwidth]{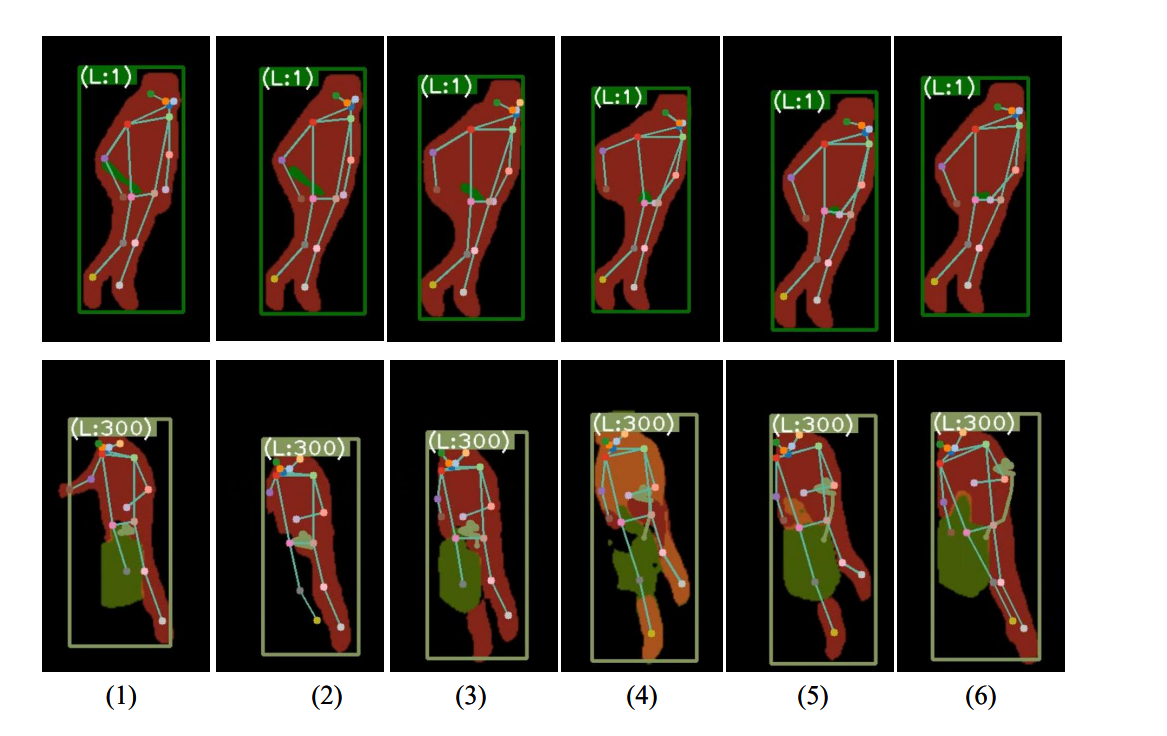}
    \caption{Two six-frame shoplifting sequences are illustrated, with the top row showing concealment of an item in a pocket and the bottom row showing placement of an item in a bag while standing. Anonymized pose sequences ensure privacy while enabling IoT deployment.}
    \label{fig:sample}
\end{figure}

\cref{fig:sample} illustrates two anonymized pose sequences corresponding to pocket concealment and bag concealment.  

\subsection{Dataset Statistics}

\begin{table*}[htp]
  \small
  \centering
\caption{Separation of training and test sets for model benchmarking. The Poselift dataset Set\cite{Rashvand_2025_WACV}  provides a baseline, while our dataset expands it by incorporating a much larger volume of real-world shopping behavior in the training set, extending the real-world test set, and adding a staged test branch. Each dataset reports the number of normal and shoplifting frames, shoplifting samples, and camera views, with test sets designed to contain nearly equal numbers of shoplifting and normal frames.}
  \resizebox{0.9\textwidth}{!}{
  \begin{tabular}{@{}c c c c c c c c@{}}
    \toprule
    \toprule
    \textbf{Dataset} & \textbf{Normal Frames} & \textbf{Shoplifting Frames} & \textbf{Shoplifting Samples} & \textbf{Camera Views} \\
    PoseLift Train Set\cite{Rashvand_2025_WACV}  & 53,353 & 0 & 0 & 6 \\
    PoseLift Test set\cite{Rashvand_2025_WACV}  & 2,221 & 1,500 & 43 & 6 \\ \midrule
    RetailS Train Set (\textbf{Ours}) & 19,971,589 & 0 & 0 & 6 \\
    RetailS Real-world Test set  (\textbf{Ours}) & 2,432 & 1,933 & 53 & 6 \\
    RetailS Staged Test Set (\textbf{Ours}) & 20,578 & 20,335 & 898 & 6 \\ 
    \bottomrule
    \bottomrule
  \end{tabular}}

  \label{tab:dataset_separation}
\end{table*}

Table~\ref{tab:dataset_separation} compares Poselift \cite{Rashvand_2025_WACV} and RetailS. RetailS is significantly larger, multi-view, and IoT-oriented, with nearly 20M normal frames captured from real shopping activities. The staged subset includes 898 incidents across five concealment categories, while the real-world subset includes 53 authentic incidents spanning two years of IoT surveillance. This design ensures domain diversity and realistic evaluation under IoT conditions.  

\subsubsection{Camera View Statistics}
\begin{figure}[htp]
  \centering
  \includegraphics[width=3.2in]{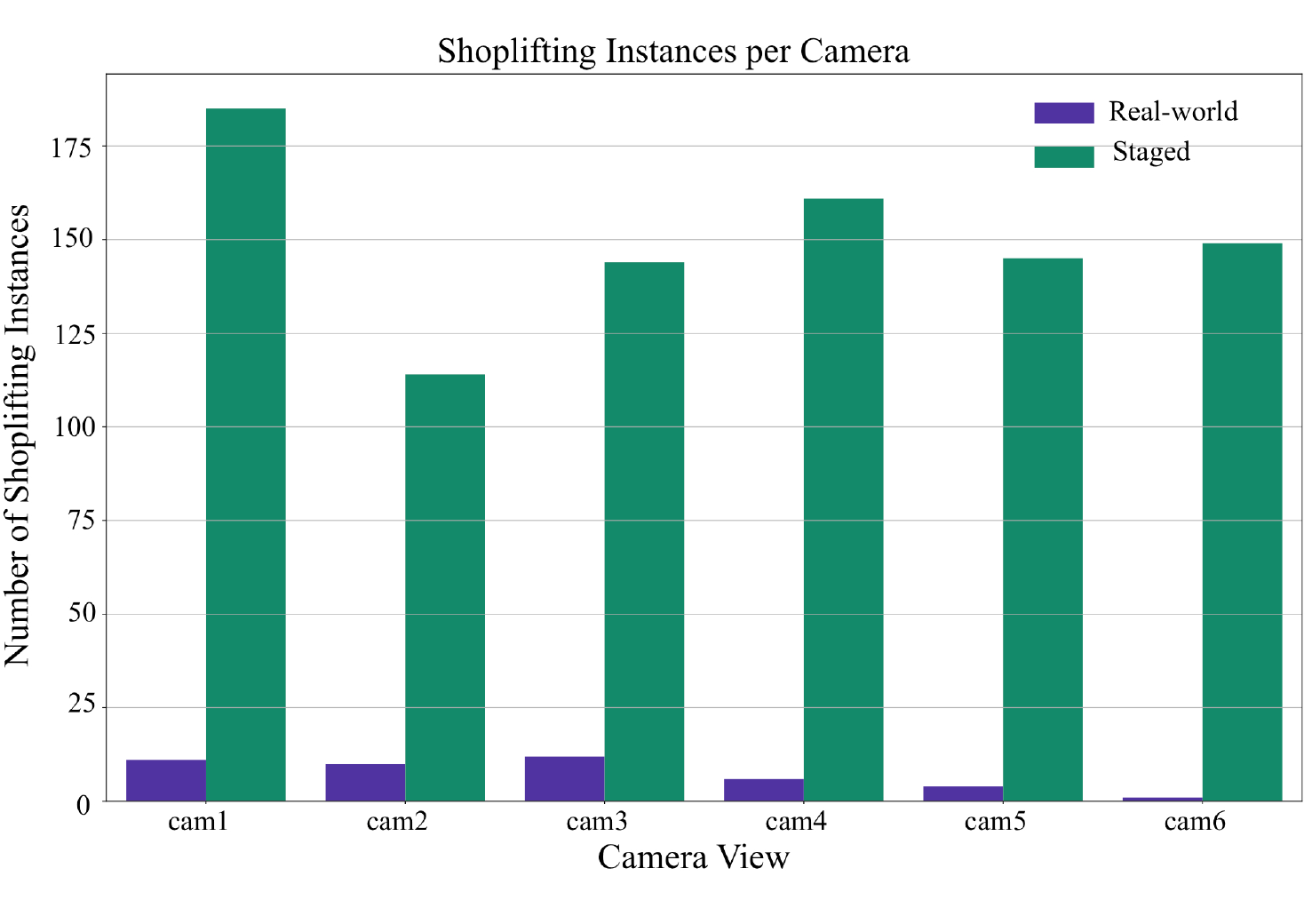}
  \caption{Distribution of shoplifting instances across six IoT-connected cameras. Real-world events cluster in specific zones, while staged scenarios ensure coverage across all views.}
  \label{fig:camera_view}
\end{figure}

IoT deployment introduces spatial heterogeneity, with incidents unevenly distributed across camera views. As shown in \cref{fig:camera_view}, real-world incidents cluster around certain store zones (e.g., cameras 1–3), while staged events are more uniformly distributed by design. This ensures evaluation covers both skewed and balanced spatial distributions, reflecting operational IoT realities.  

\subsubsection{Distribution of Shoplifting Categories}
\begin{figure}[htp]
  \centering
  \includegraphics[width=3.2in]{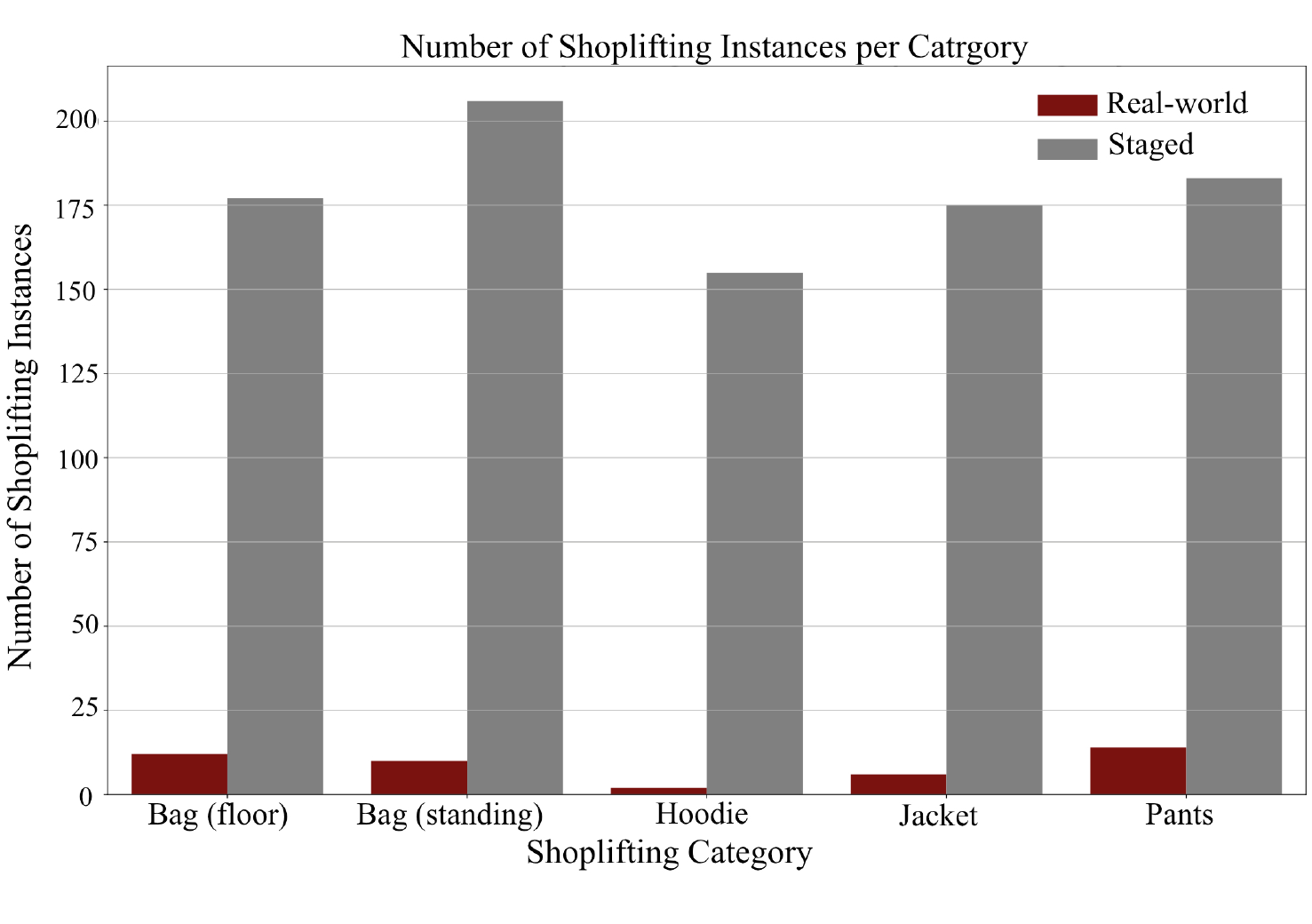}
  \caption{Distribution of shoplifting strategies in RetailS across staged and real-world subsets. Balanced staged coverage supports generalization in IoT deployments.}
  \label{fig:dis_categories}
\end{figure}

\cref{fig:dis_categories} compares concealment strategies across subsets. Real-world incidents are skewed (e.g., hiding in pants most frequent), while staged events are balanced across categories, ensuring fair evaluation. Such balance is crucial for IoT anomaly detectors to generalize beyond biased real-world distributions.  

These observations demonstrate that RetailS was designed not only for benchmark performance but also to reflect the constraints and variability of IoT-driven retail deployment. By combining scale, diversity, privacy-preserving representation, and multi-camera coverage, RetailS bridges the gap between controlled lab datasets and the operational realities of IoT-enabled retail environments.
\section{Experiment and Evaluation}

\noindent\textbf{Motivation for IoT-Oriented Evaluation.}
Unlike conventional anomaly detection benchmarks, our goal is not only to measure accuracy but also to emulate the execution cycle of a real-world IoT retail deployment. This motivates several design choices. First, thresholds for anomaly filtering are fixed once (from offline calibration) and then held constant, reflecting the operational stability needed for edge devices. Second, model weights are periodically refreshed using pseudo-labeled buffers, emulating scheduled back-end adaptation under drift. Third, data is divided into time-sliced windows (half-day and full-day), mirroring store operational cycles and realistic maintenance intervals. These settings ensure that our experiments test both algorithmic effectiveness and the feasibility of periodic adaptation under IoT resource and latency constraints.

We evaluate three state-of-the-art pose-based anomaly detection models: STG-NF~\cite{hirschorn2023normalizing}, SPARTA~\cite{noghre2024human}, and TSGAD~\cite{noghre2024exploratory}. Models are trained and validated in two settings: (i) an \textbf{offline baseline} that trains once on normal-only data, establishing a static reference, and (ii) a \textbf{periodic adaptation} protocol (Sec.~\ref{sec:method}) that updates models incrementally with pseudo-labeled, time-sliced batches to test adaptation under domain shift. Evaluation is performed using ranking metrics (AUC-ROC/AUC-PR) as well as fixed-threshold operating points, $\mathrm{F1}@\text{thr}_{F1}$ and $H_{\text{PRS}}@\text{thr}_{H_{\text{PRS}}}$, the latter explicitly penalizing false alarms, which is critical in IoT deployments. All experiments were run on a server with dual 64-core CPUs, 512\,GB memory, and three A6000 GPUs (256\,GB total), but the pipeline stages are structured to model edge–cloud division typical of IoT systems.

\begin{table}[htp]
\centering
\caption{Performance of three state-of-the-art anomaly detection models under offline training. PoseLift rows show models trained on PoseLift and tested on the RetailS, while RetailS rows use our training set, highlighting generalization and the effect of larger, more diverse data.}
\label{tab:Offline_model_performance}
\resizebox{\columnwidth}{!}{
\begin{tabular}{cccccc}
\cline{1-2}
\toprule
\toprule
\multirow{2}{*}{\textbf{Dataset}} & \multirow{2}{*}{\textbf{Model}} & \multicolumn{2}{c}{\textbf{Staged Test Set}} & \multicolumn{2}{c}{\textbf{Real-world Test Set}} \\ \cline{3-6} 
 &  & AUC-ROC & AUC-PR & AUC-ROC & AUC-PR \\ \midrule
\multirow{3}{*}{PoseLift} & STG-NF \cite{hirschorn2023normalizing} & \textbf{88.10} & 86.46 & 61.35 & 33.72 \\
 & TSGAD\cite{noghre2024exploratory} & \textbf{59.37} & 40.83 & 59.08 & 35.94 \\
 & SPARTA \cite{noghre2024human} & \textbf{77.84} & \textbf{75.81} & 56.82 & 34.42 \\ \midrule
\multirow{3}{*}{RetailS (\textbf{Ours})} & STG-NF \cite{hirschorn2023normalizing} & 87.24 & \textbf{86.60} & \textbf{63.22} & \textbf{38.44} \\
 & TSGAD \cite{noghre2024exploratory} & 51.99 & \textbf{51.87} & \textbf{62.16} & \textbf{38.32} \\
 & SPARTA \cite{noghre2024human} & 74.93 & 72.42 & \textbf{58.23} & \textbf{34.97} \\
\bottomrule
\bottomrule
\end{tabular}}
\label{offline_benchmarking}
\end{table}

\begin{figure*}[htbp]
  \centering
  \begin{subfigure}{0.45\textwidth}
    \includegraphics[width=\linewidth, trim=12 23 15 74, clip]{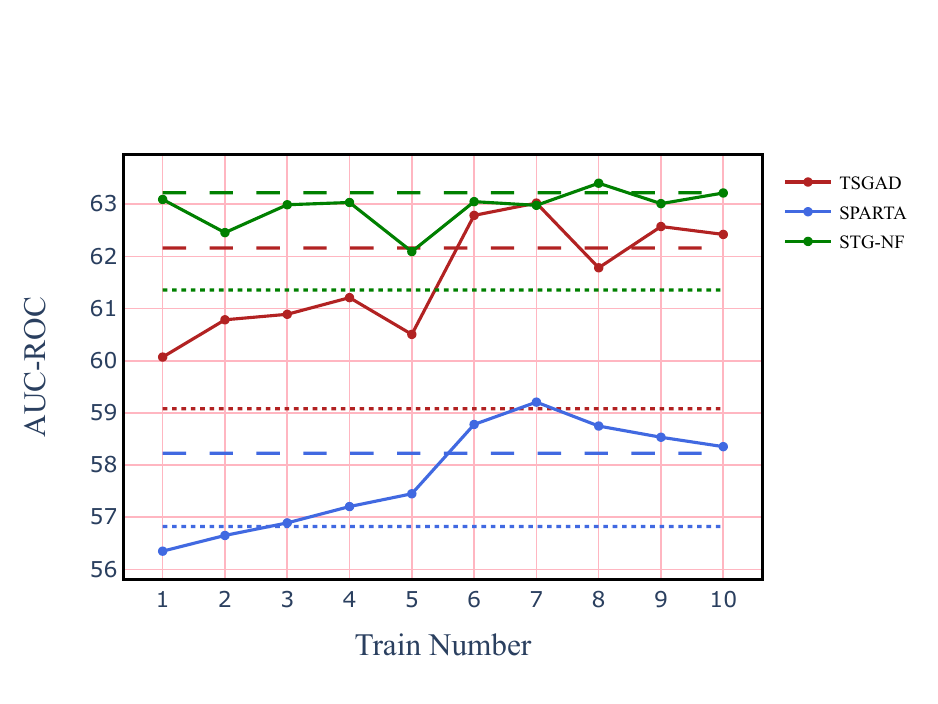}
    \caption{F1 score threshold}
  \end{subfigure}
  \hfill
  \begin{subfigure}{0.45\textwidth}
    \includegraphics[width=\linewidth, trim=12 23 15 74, clip]{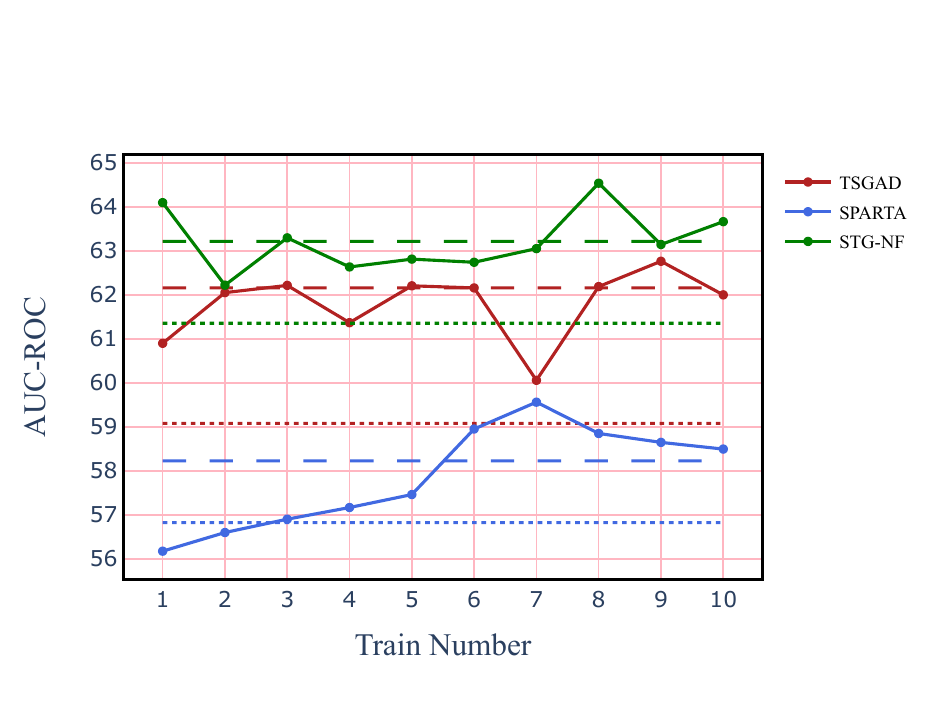}
    \caption{$H_{\text{PRS}}$ score threshold}
  \end{subfigure}
  \medskip
  \begin{subfigure}{0.45\textwidth}
    \includegraphics[width=\linewidth, trim=12 23 15 74, clip]{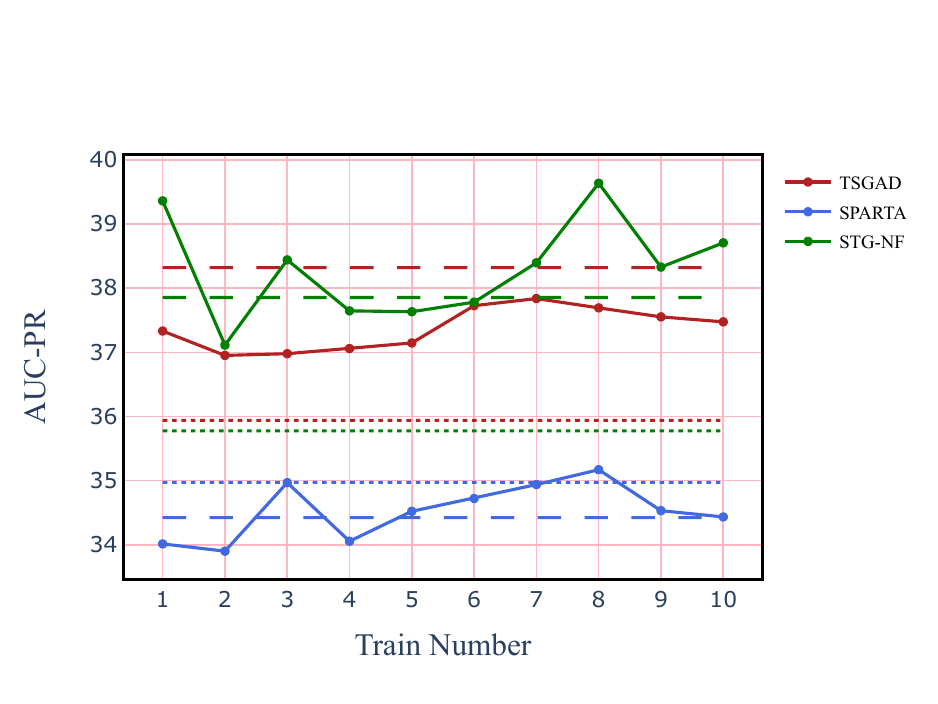}
    \caption{F1 score threshold}
  \end{subfigure}
  \hfill
  \begin{subfigure}{0.45\textwidth}
    \includegraphics[width=\linewidth, trim=12 23 15 74, clip]{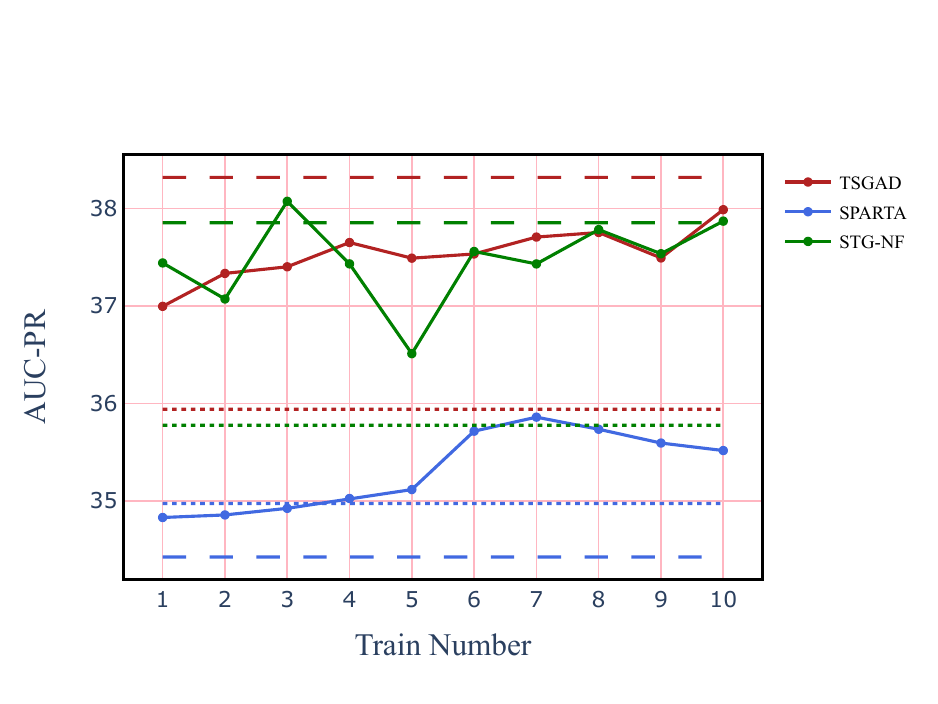}
    \caption{$H_{\text{PRS}}$ score threshold}
  \end{subfigure}
  \caption{Model performance trends at \textbf{one-day update frequency}. Long dashes = offline training on RetailS, dots = offline training on PoseLift, solid lines = periodic adaptation.}
  \label{fig:oneday}
\end{figure*}

\begin{figure*}[htbp]
  \centering
  \begin{subfigure}{0.45\textwidth}
    \includegraphics[width=\linewidth, trim=12 23 15 74, clip]{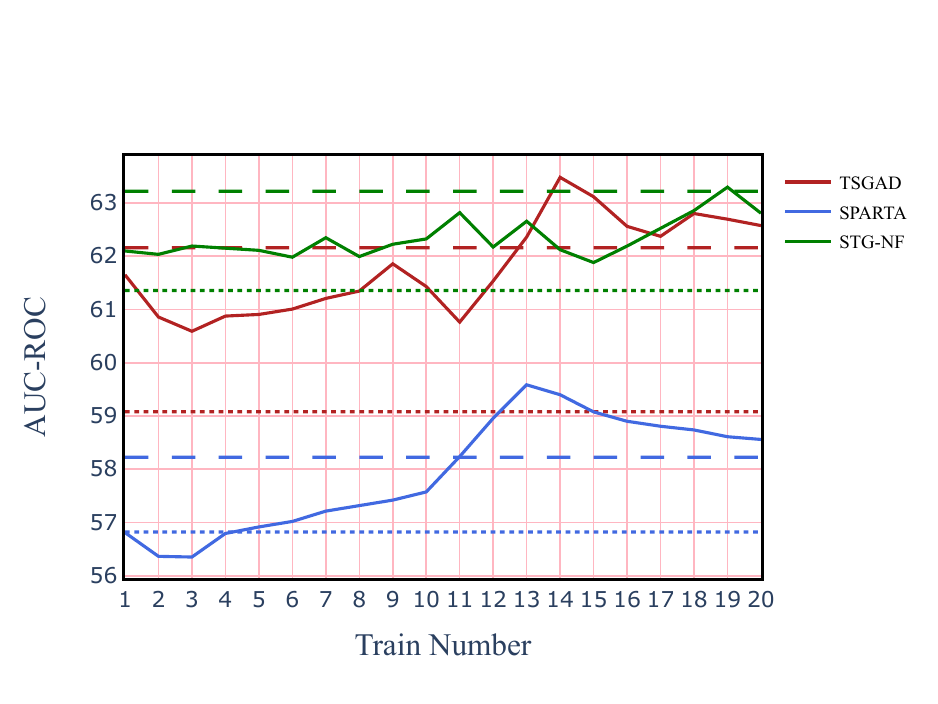}
    \caption{F1 score threshold}
  \end{subfigure}
  \hfill
  \begin{subfigure}{0.45\textwidth}
    \includegraphics[width=\linewidth, trim=12 23 15 74, clip]{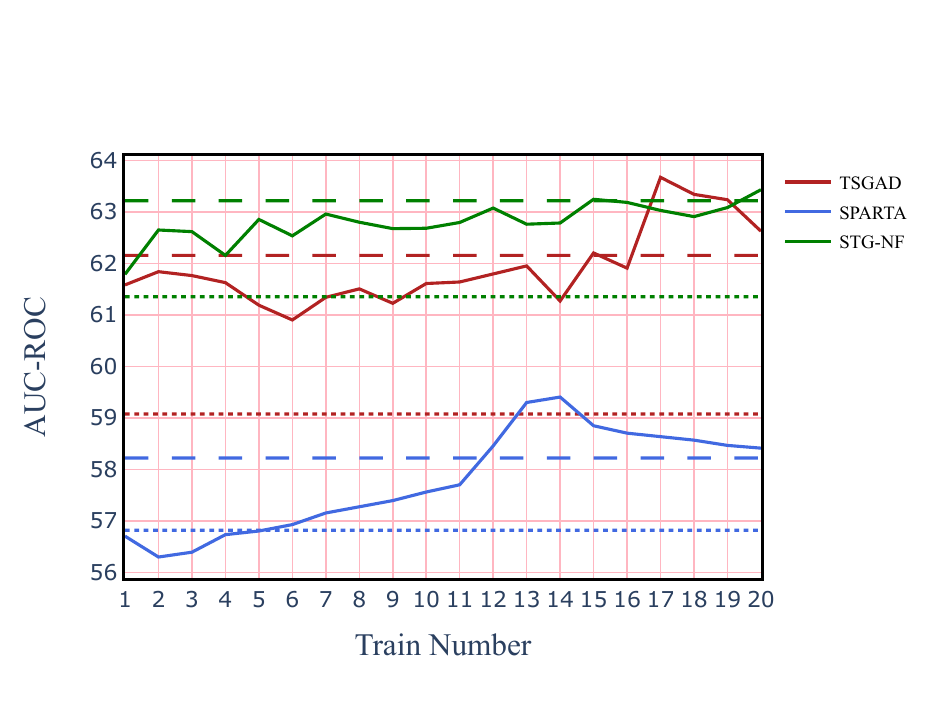}
    \caption{$H_{\text{PRS}}$ score threshold}
  \end{subfigure}
  \medskip
  \begin{subfigure}{0.45\textwidth}
    \includegraphics[width=\linewidth, trim=12 23 15 74, clip]{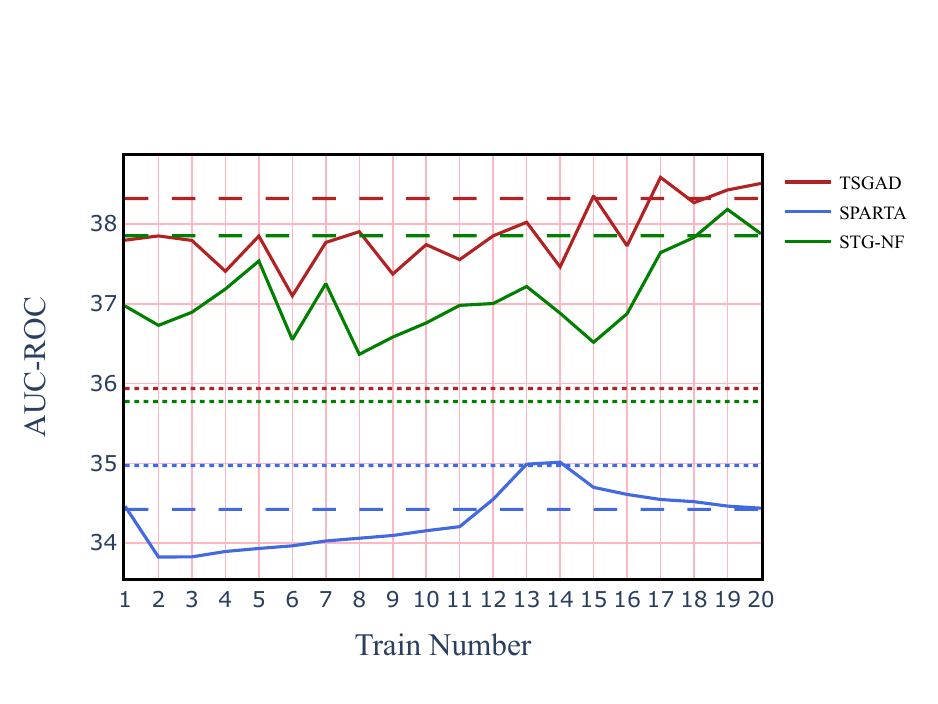}
    \caption{F1 score threshold}
  \end{subfigure}
  \hfill
  \begin{subfigure}{0.45\textwidth}
    \includegraphics[width=\linewidth, trim=12 23 15 74, clip]{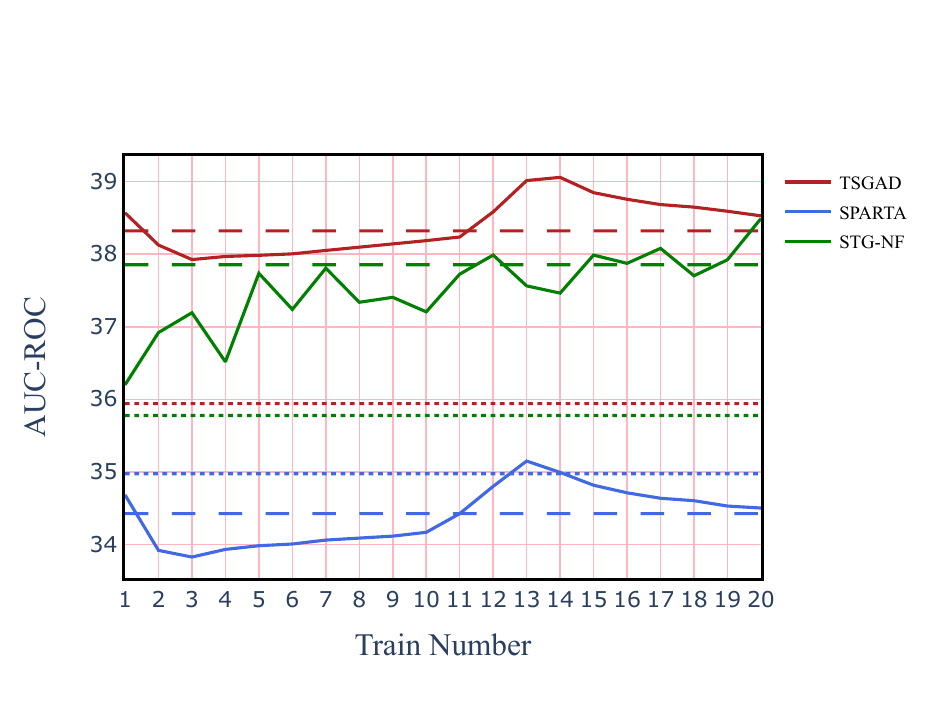}
    \caption{$H_{\text{PRS}}$ score threshold}
  \end{subfigure}
  \caption{Model performance trends at \textbf{half-day update frequency}. Long dashes = offline training on RetailS, dots = offline training on PoseLift, solid lines = periodic adaptation.}
  \label{fig:halfday}
\end{figure*}

\subsection{Offline Training Benchmark}
We begin with the offline training paradigm, the dominant approach in video anomaly detection \cite{hirschorn2023normalizing, noghre2024human,markovitz2020graph,rashvand2025shopformer}. Models are trained once on normal data and tested on evaluation sets with both normal and shoplifting behavior.  

\cref{tab:Offline_model_performance} reports results on staged and real-world test sets, trained on PoseLift vs.\ RetailS. Two insights emerge. First, staged results consistently exceed real-world ones (e.g., STG-NF achieves $\mathrm{AUC\text{-}ROC}=88.10$/$\mathrm{AUC\text{-}PR}=86.46$ staged, but only $61.35$/$33.72$ real-world), showing the impact of drift in cameras, layout, and shopper behavior. Staged thus approximates a best-case, while real-world reflects the deployment baseline under IoT drift. Second, training on the more diverse RetailS set improves generalization (e.g., STG-NF rises to $63.22$ AUC-ROC / $38.44$ AUC-PR on real-world), demonstrating the dataset’s value for realistic IoT adaptation.

\subsection{periodic adaptation Benchmark}
As outlined in Sec.~\ref{sec:method}, periodic adaptation was tested with staged anomalies injected into normal streams, while real-world test data remained unseen. Crucially, thresholds ($\text{thr}_{F1}$, $\text{thr}_{H_{\text{PRS}}}$) were fixed from the offline calibration split, emulating IoT edge devices operating at stable decision points. Model weights, however, were periodically refreshed from time-sliced buffers, reflecting the edge–back-end split in IoT execution.

\cref{fig:oneday,fig:halfday} trace performance under daily and half-day update frequencies. Three observations stand out: (i) periodic adaptation outperforms offline baselines in 91.6\% of evaluations, confirming the benefit of adaptation under drift; (ii) $H_{\text{PRS}}$ thresholds outperform F1 in 9/12 cases, showing the value of explicitly controlling false alarms in IoT deployments; and (iii) half-day updates outperform daily updates, suggesting fresher buffers capture local drift better—supporting more frequent adaptation when compute budgets allow.

\subsection{Periodic Adaptation Time Consumption}
To stress-test practicality, we implemented preprocessing following Pazho et al.~\cite{pazho2023ancilia} and Yao et al.~\cite{yao2025lab}, running six Ancilia AI pipelines concurrently. Throughput reached $\approx 26$ FPS with end-to-end latency $\approx 20$\,s, showing viability under multi-stream IoT conditions.  \cref{tab:time} shows update times: SPARTA~\cite{noghre2024human} requires 2–3 minutes, STG-NF~\cite{hirschorn2023normalizing} 3.5–7.3 minutes, and TSGAD~\cite{noghre2024exploratory} up to 27 minutes (half-day) or $>$1 hour (daily). All updates complete within their scheduling windows, ensuring timely adaptation. However, heavier models like TSGAD may be impractical for frequent IoT refresh, while lighter ones suit real-time adaptation.

\begin{table}[htp]
\label{tab:time}
\centering
\caption{Average training time (in minutes) per update for continual learning with half-day and one-day data batches across three state-of-the-art pose-based models.}
\label{tab:time}
\begin{tabular}{c|cc}
\toprule
\toprule
\multirow{2}{*}{Model} & \multicolumn{2}{c}{avg. time  (min.)} \\ 
                       & Half-day data      & One-day data     \\ \midrule
STG-NF\cite{hirschorn2023normalizing}                 & 3.5                & 7.3              \\
TSGAD\cite{noghre2024exploratory}                  & 26.8               & 65               \\
SPARTA\cite{noghre2024human}                 & 2.05               & 3.2              \\
\bottomrule
\bottomrule
\end{tabular}
\end{table}

\subsection{Ablation and Deployment Insights}

To examine the role of thresholding strategies in periodic adaptation, we conducted an ablation contrasting fixed thresholds (determined once during offline calibration) with adaptive thresholds (recalibrated after each update). While the adaptive approach yielded modest improvements of approximately 1–2\% in AUC-based metrics, it introduced instability, with false positive rates fluctuating considerably across windows. In contrast, fixed thresholds demonstrated more stable and predictable operation, supporting the IoT-motivated principle of decoupling inference at the edge (which benefits from stable decision rules) from adaptation in the back-end (which updates model weights periodically). Building on this analysis, and synthesizing the results presented in Tables~\ref{tab:Offline_model_performance}, \ref{tab:time}, and Figures~\ref{fig:oneday}–\ref{fig:halfday}, we distill the following deployment insights:  

\begin{itemize}
\item \textbf{Thresholding Policy:} As shown in Figures~\ref{fig:oneday} and \ref{fig:halfday}, thresholds selected using $H_{\text{PRS}}$ outperform F1-based thresholds in 9 out of 12 evaluations, yielding more reliable false-alarm control in IoT retail settings. In our deployment scenario, thresholds are calibrated once on an offline validation split and kept fixed during inference, reflecting practical constraints where frequent recalibration is costly. To evaluate robustness under domain drift, we analyze periodic retraining over daily and half-day windows (Figs.~8–9), holding thresholds fixed for up to 10 daily and 20 half-day updates. Both AUC-ROC and AUC-PR remain stable over time, while periodic retraining consistently improves performance without threshold adjustment.

  \item \textbf{Update Frequency:} Performance trends in Figures~\ref{fig:oneday} and \ref{fig:halfday} demonstrate that half-day update intervals yield higher AUC-ROC and AUC-PR scores than daily updates, confirming that shorter adaptation cycles better capture domain drift. Daily updates remain a viable lower-cost fallback when computational or bandwidth resources are constrained.  
  \item \textbf{Model Selection:} Time consumption results in Table~\ref{tab:time} reveal that lightweight models such as SPARTA and STG-NF complete updates in under 10 minutes, making them suitable for frequent IoT adaptation. In contrast, TSGAD requires up to 27 minutes (half-day) or over an hour (daily), making it less practical for real-time deployment.  
  \item \textbf{Latency and Resource Budgeting:} As indicated by throughput and latency measurements in Table~\ref{tab:time}, continual updates can be completed well within their respective scheduling windows for lightweight models, ensuring timeliness of adaptation and preventing model staleness under multi-stream IoT conditions.  
  \item \textbf{Edge–Cloud Split:} Combining the stability of fixed thresholds (as confirmed by the ablation study) with periodic back-end weight updates (Figures~\ref{fig:oneday}–\ref{fig:halfday}) provides the optimal trade-off between inference stability at the edge and adaptability in the back-end, reflecting the operational requirements of IoT deployments.  
\end{itemize}

\section{Conclusion}
We presented a privacy-preserving, pose-based framework for shoplifting detection and demonstrated how periodic adaptation closes the gap between offline benchmarks and IoT-enabled real-world deployment. By introducing \textit{RetailS}—a large-scale dataset spanning millions of normal frames and both staged and authentic shoplifting events—we established strong offline baselines and showed how performance degrades under domain shift, reflecting the realities of IoT surveillance environments.  

Our findings highlight a central lesson: offline training alone is insufficient for real-world IoT deployment. Static models fail to keep pace with changing store layouts, camera placements, lighting conditions, and evolving shopper behavior. In contrast, periodic adaptation with pseudo-labeled buffers enables systems to adapt to drift over time. Critically, fixing thresholds once offline and updating only model weights mirrors the operational split in IoT systems between stable edge inference and scheduled back-end updates. This strategy not only improves detection accuracy but also provides more stable, generalizable decision thresholds.  From our evaluation, two key deployment insights emerge. First, thresholds selected with $H_{\text{PRS}}$ (with F1 as a companion) provide better control of false alarms, reducing nuisance alerts in IoT environments. Second, more frequent updates (half-day) outperform daily updates by keeping pace with drift while still respecting compute budgets.

\section*{Acknowledgments}

This research is supported by the National Science Foundation (NSF) under Award No. 1831795.

\bibliographystyle{ieeetr}  
\bibliography{main} 

\clearpage
\setcounter{page}{1}

\section*{Supplementary Material}
\label{sec:sup}

This supplementary material provides additional analyses and clarifications that support the main findings of the paper, with particular emphasis on deployment-related variability and robustness under real-world IoT conditions. 
Our experimental setting does not assume clean or ideal camera streams. On the contrary, both our dataset design and input representation inherently reflect noisy and incomplete measurements commonly observed in deployed retail surveillance systems.
First, all data used in this study is collected from a live, operational retail store, using existing in-store CCTV infrastructure under normal business conditions (Sec. IV). As illustrated by the store layout and multi-camera configuration (Fig.3),  the environment naturally contains occlusions, partial views, crowd interference, and viewpoint inconsistencies, which frequently result in missing or unstable pose keypoints.
Second, our framework operates on pose keypoints rather than RGB pixels. The conversion from RGB video to pose sequences (via detection, tracking, and pose estimation algorithms) introduces an additional layer of realistic noise, including tracking errors, and missing pose keypoints. This is explicitly addressed in our data preparation pipeline, where missing keypoints are interpolated and temporal smoothing is applied to stabilize extracted keypoints under streaming IoT conditions (Sec. IV-B). As a result, the models are trained and evaluated on pose data that already contains noise and partial observations.
While we do not inject synthetic camera failures explicitly, our evaluation reflects robustness under naturally occurring noise in real-world IoT deployments. This design enables the system to adapt to gradual camera degradation, layout changes, and evolving occlusion patterns without relying on clean data assumptions.

\begin{figure}[htp]
  \centering
  \includegraphics[width=3.2in]{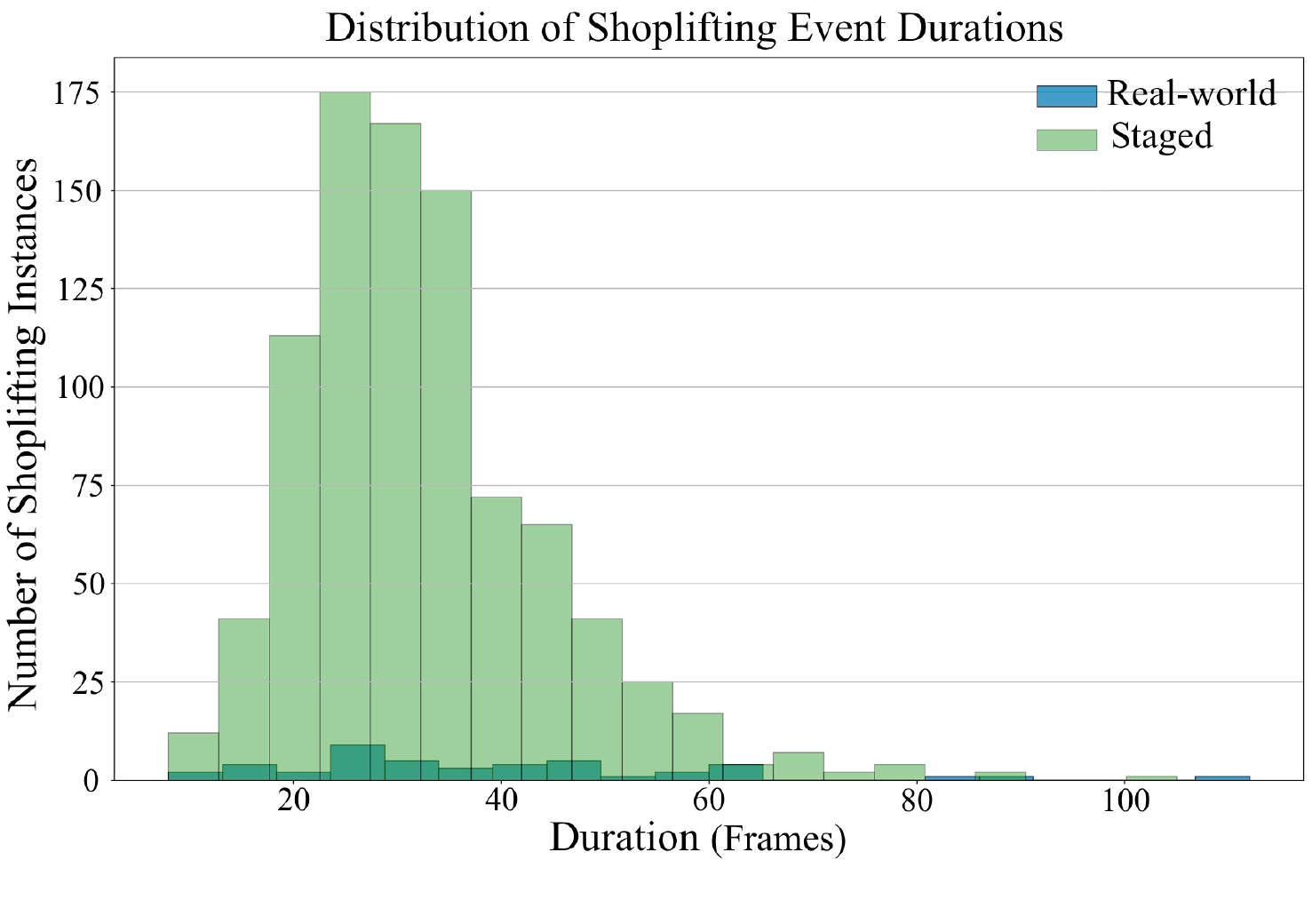}
  \caption{Distribution of shoplifting event durations in staged and real-world subsets. Short durations reflect realistic concealment strategies observed in IoT surveillance.}
  \label{fig:his}
\end{figure}

\cref{fig:his} shows histograms of shoplifting event durations. Both staged and real-world distributions are concentrated between 25–35 frames, consistent with brief concealment behaviors observed in live IoT surveillance. Unlike earlier staged datasets, these events were not artificially lengthened, reflecting realistic adversarial timing.  

Our evaluation provides a comprehensive analysis of performance variability across the sources of variance in real-world retail IoT deployments, including camera-view variability, shoplifting behavior variability, and temporal drift.

First of all, to quantify the effect of camera viewpoint on detection performance, we conducted a per-camera analysis in which each model is trained using the Retails training subset corresponding to that camera and then evaluated on the corresponding staged per-camera test split. This experimental setup isolates the influence of viewpoint, occlusion patterns, and camera placement while keeping all other factors fixed. As shown in \cref{tab:per_camera_performance}, shoplifting incidents are unevenly distributed across the six cameras, and the resulting per-camera metrics exhibit noticeable differences. For example, STG-NF achieves an AUC-ROC of 83.46 on C4 but drops to 70.68 on C6, a difference of 12.78\%. SPARTA shows a 7.5 percentage point spread (from 71.99 on C4 to 79.51 on C2), while TS-GAD varies by 2.91 points (from 51.22 on C6 to 54.13 on C1). Comparing models trained on all camera views (\cref {tab:Offline_model_performance}) against those trained and evaluated on a single camera (\cref{tab:per_camera_performance}) reveals the importance of viewpoint diversity. For example, for STG-NF, AUC-ROC drops from 87.24 under multi-camera training to an average of 77.77 under single-camera training. SPARTA exhibits a more moderate viewpoint sensitivity, with a spread of 7.5\% across cameras, while TS-GAD shows relatively small variation but tends to overfit to individual views rather than generalize across them. 
These differences indicate that camera-specific characteristics constitute a significant source of performance degradation. Overall, the per-camera evaluation demonstrates that viewpoint drift is a major contributor to domain variation in this setting. This provides strong motivation for multi-view periodic retraining when deploying models across newly reconfigublue camera views in practical retail environments.

\begin{table}[htpb]
\centering
\captionsetup{font={color=black}}
\caption{Per-camera performance comparison under single-camera training and evaluation. Each model is trained using only normal data from one camera view and evaluated on the corresponding per-camera test split. Results highlight the impact of viewpoint, occlusion patterns, and camera placement on detection performance.}
\label{tab:per_camera_performance}
\resizebox{\columnwidth}{!}{\color{black}
\begin{tabular}{c c cc cc cc}
\toprule
\multirow{2}{*}{\textbf{Camera}} &
\multirow{2}{*}{\textbf{Shoplifting Events}} &
\multicolumn{2}{c}{\textbf{STG-NF}} &
\multicolumn{2}{c}{\textbf{TS-GAD}} &
\multicolumn{2}{c}{\textbf{SPARTA}} \\
\cmidrule(lr){3-4} \cmidrule(lr){5-6} \cmidrule(lr){7-8}
 &  & AUC-ROC & AUC-PR & AUC-ROC & AUC-PR & AUC-ROC & AUC-PR \\
\midrule
C1 & 185 & 78.15 & 73.58 & 54.13 & 57.70 & 75.25 & 71.65 \\
C2 & 114 & 76.17 & 75.33 & 53.07 & 52.42 & 79.51 & 77.61 \\
C3 & 144 & 76.40 & 73.46 & 52.39 & 54.09 & 76.92 & 74.64 \\
C4 & 161 & 83.46 & 79.26 & 52.06 & 53.43 & 71.99 & 69.90 \\
C5 & 145 & 79.76 & 77.10 & 52.46 & 53.87 & 72.60 & 69.85 \\
C6 & 149 & 70.68 & 68.49 & 51.22 & 51.56 & 78.24 & 76.07 \\
\bottomrule
\end{tabular}}
\end{table}

\cref {tab:per_event_performance} compares per-shoplifting-event detection performance across three pose-based models under all camera training and per-event evaluation. Overall, hoodie and jacket-based concealment events achieve the strongest performance across all models, with consistently high AUC-ROC and AUC-PR scores. These events typically involve clear upper-body articulation and sustained concealment motions, which are well captublue by pose dynamics.
In contrast, bag (standing) events exhibit the weakest performance, particularly in AUC-PR, indicating higher false-alarm sensitivity. This behavior is often subtle, short in duration, and visually similar to normal shopping actions, making it more difficult to distinguish from benign motion patterns using pose alone. Pants-based concealment shows moderate performance, falling between jacket/bag (floor) events and bag (standing), suggesting that lower-body occlusions and limited joint visibility blueuce discriminative motion cues. Overall, all the performance shows really close and average distribution across all the events with different models without much strong disadvantage.
Across event types, STG-NF consistently achieves the highest or near-highest scores, especially on more challenging cases such as bag (standing) and pants, showing stronger robustness to subtle motion variations. These results highlight that event-specific motion characteristics strongly influence detection difficulty, motivating both event-diverse evaluation and adaptive deployment strategies in real-world retail settings.

\begin{table}[htpb]
\centering
\captionsetup{font={color=black}}
\caption{Per-shoplifting event performance comparison under all-camera training and evaluation. Each model is trained using only normal data and evaluated on the corresponding per-shoplifting event test split. Results highlight the average performace ofdifferent shoplifting event in RetailS dataset.}
\label{tab:per_event_performance}
\resizebox{\columnwidth}{!}{\color{black}
\begin{tabular}{c cc cc cc}
\toprule
\multirow{2}{*}{\textbf{Event}} &
\multicolumn{2}{c}{\textbf{STG-NF}} &
\multicolumn{2}{c}{\textbf{TS-GAD}} &
\multicolumn{2}{c}{\textbf{SPARTA}} \\
\cmidrule(lr){2-3} \cmidrule(lr){4-5} \cmidrule(lr){6-7}
  & AUC-ROC & AUC-PR & AUC-ROC & AUC-PR & AUC-ROC & AUC-PR \\
\midrule
Bag (floor)    & 86.72 & 86.52 & 51.85 & 54.59 & 74.28 & 74.52 \\
Bag (standing) & 86.11 & 86.31  & 52.09 & 53.26 & 74.11 & 75.02 \\
Hoodie & 87.99 & 87.43 & 52.52 & 52.99 & 74.83 & 76.58 \\
Jacket & 87.96 & 86.59 & 52.72 & 52.25 & 74.76 & 77.53 \\
Pants  & 87.90 & 86.78 & 51.65 &  51.59 & 74.87 & 76.47 \\
\bottomrule
\end{tabular}}
\end{table}

Overall, as reported in the supplementary Material, we explicitly analyze per-camera performance variance under single-camera training and evaluation (Tab. IV), revealing noticeable performance differences caused by camera viewpoint, occlusion patterns, and camera placement. For example, STG-NF has up to 13\% AUC-ROC variation across different cameras. We further analyzed per-event performance variance across different shoplifting concealment strategies (Tab. V), where performance differs across behaviors. This captures behavioral uncertainty inherent in real world incidents rather than random noise, and directly reflects the diversity of anomalous patterns encounteblue in practice. 
In addition to spatial and behavioral variability, our periodic adaptation protocol inherently produces repeated evaluations over time, with 10 daily and 20 half-day update cycles. It provides repeated measurements across independent time windows rather than a single static evaluation. We further clarify that all experiments are conducted with fixed random seeds and deterministic training settings, ensuring that all reported results are fully reproducible.

\begin{algorithm}
\caption{Periodic Adaptation Algorithm}
\label{alg:periodic_adapt_v3}
\begin{algorithmic}[1]

\STATE \textbf{Initialization:}
\STATE \quad Load the current VAD model weights $W^{(0)}$ and set anomaly threshold $\tau$.
\STATE \quad Set window size $L=24$ frames.
\STATE \quad Initialize a low-score buffer $\mathcal{D}_{\text{low}} \leftarrow \emptyset$ and an adaptation timer.
\STATE \quad Prepare an abnormal pool $\mathcal{A}$ as a subset sampled from the existing test data.
\STATE \quad Set the normal-to-abnormal mixing ratio to $9{:}1$.

\FOR{\textbf{each} camera dataset stream $\mathcal{S}_{c_i}$ \textbf{in} the set of cameras}
  \WHILE{\textit{data stream from} $\mathcal{S}_{c_i}$ \textit{is available}}

    \STATE \textbf{Step 1: Batch Assembly (Dataset Input)}
    \STATE \quad Fetch a batch $\mathcal{B}$ of length $L=24$ from $\mathcal{S}_{c_i}$.
    \STATE \quad $\mathcal{B}$ contains tuples $(\textit{frame\_id}, \textit{person\_id}, \textit{kpts})$.

    \STATE \textbf{Step 2: Normal/Abnormal Mixing}
    \STATE \quad Sample normal windows from $\mathcal{B}$ and abnormal continuous windows from $\mathcal{A}$.
    \STATE \quad Compose a mixed batch $\tilde{\mathcal{B}}$ with a $9{:}1$ normal-to-abnormal ratio.

    \STATE \textbf{Step 3: VAD Inference and Thresholding}
    \STATE \quad Send $\tilde{\mathcal{B}}$ to the VAD model parameterized by current weights $W^{(t)}$.
    \STATE \quad Obtain anomaly score $s$ for the batch.

    \STATE \textbf{Step 4: Low-Score Data Collection}
    \IF{$s < \tau$}
      \STATE \quad Append $\tilde{\mathcal{B}}$ (and metadata) to the low-score buffer $\mathcal{D}_{\text{low}}$.
    \ELSE
      \STATE \quad Continue streaming without buffering.
    \ENDIF

    \STATE \textbf{Step 5: Periodic Adaptation Trigger (Runs in Parallel)}
    \IF{\textit{elapsed time} $\geq 12\text{h}$ \textbf{or} $\geq 24\text{h}$}
      \STATE \quad Launch an asynchronous training job using $\mathcal{D}_{\text{low}}$ and current weights $W^{(t)}$.
      \STATE \quad Swap buffers to keep collection running: $\mathcal{D}_{\text{train}} \leftarrow \mathcal{D}_{\text{low}}$, then $\mathcal{D}_{\text{low}} \leftarrow \emptyset$.
      \STATE \quad Reset the adaptation timer while inference and buffering continue.
    \ENDIF

    \STATE \textbf{Step 6: Weight Update (After Training Completion)}
    \IF{\textit{training job finished}}
      \STATE \quad Receive updated weights $W^{(t+1)}$ from the server.
      \STATE \quad Atomically update the deployed model: $W^{(t)} \leftarrow W^{(t+1)}$.
      \STATE \quad Continue streaming inference and filling $\mathcal{D}_{\text{low}}$ in parallel.
    \ENDIF

  \ENDWHILE
\ENDFOR

\end{algorithmic}
\end{algorithm}

\cref{alg:periodic_adapt_v3} formalizes the periodic adaptation mechanism of our video anomaly detection (VAD) system under continuous deployment. Instead of operating on raw video streams, the algorithm processes pre-extracted motion data from each camera, organized as fixed-length windows of 24 frames containing frame indices, person identities, and skeletal keypoints. Each incoming window is optionally mixed with a small proportion of continuous abnormal motion segments sampled from an existing test subset, forming a combined batch that reflects a realistic normal–abnormal ratio. The batch is then forwarded to the VAD model to obtain an anomaly score via threshold-based inference. Motion windows assigned low anomaly scores are treated as high-confidence normal samples and accumulated in a dedicated low-score buffer. This buffer serves as a self-supervised data source to capture evolving normal behavior patterns over time.

Adaptation is triggered periodically based on elapsed time (12 or 24 hours) rather than per-sample updates. When the trigger condition is met, an asynchronous batch training process is launched using the accumulated low-score buffer while inference and data collection continue in parallel. Upon completion of training, the updated model parameters are atomically swapped into the running system, ensuring uninterrupted operation. The algorithm explicitly decouples inference, data collection, and adaptation, providing a clear and reproducible procedure for online deployment. This step-by-step formulation clarifies how adaptive behavior is realized in practice while maintaining a non-blocking, scalable VAD pipeline.

\end{document}